\documentclass[conference,10pt]{IEEEtran}

\usepackage{amsmath,amssymb,amsfonts}
\usepackage{graphicx}
\usepackage{cite}
\usepackage{booktabs}
\usepackage{algorithm2e}
\usepackage{xcolor}
\usepackage{hyperref}
\usepackage{subfig}
\usepackage{lipsum}
\usepackage{multirow}
\usepackage{tikz}
\usetikzlibrary{positioning, fit, arrows.meta, calc}
\hypersetup{
  colorlinks=true,
  linkcolor=black,
  citecolor=black,
  urlcolor=blue,
}

\begin{document}

\title{A Composable AI-Accelerated Iterative Solver for 3D-IC Thermal Modeling}

% Required for \thanks in IEEEtran conference mode
\IEEEoverridecommandlockouts

\author{
  \IEEEauthorblockN{
    Yixing Li, Jiahang Zhou, Zhiyu Zeng, Xin Ai
  }
  \thanks{
    Yixing Li, Zhiyu Zeng, and Xin Ai are with Cadence Design Systems
    (email: \texttt{yixingli@cadence.com};
    \texttt{zzeng@cadence.com};
    \texttt{nathanai@cadence.com}).
    Jiahang Zhou is with the University of Notre Dame
    (email: \texttt{jzhou8@nd.edu}).
  }
}
\maketitle

\begin{abstract}
Accurate thermal analysis of heterogeneous 2.5D/3D-IC packages is essential
yet computationally prohibitive.
A single full-package FEM simulation can take hours, while AI-based surrogates
treat the entire stack as a monolithic prediction target and must be retrained
whenever the die count or topology changes.
To address this limitation, this work proposes
\textbf{\underline{D}omain-Decomposed \underline{A}I-Accelerated \underline{I}terative 
\underline{S}olver for \underline{T}hermal Analysis (DAIST)},
a composable thermal solver that decomposes the global package simulation into
block-level subdomain problems, replaces subdomain solvers with neural operators,
and couples them through iterative exchanges of interfacial temperature and
heat flux.
This local-to-global architecture eliminates the topology lock-in of monolithic
models: block-level neural operators can be directly reused in unseen package
assemblies without retraining.
The iterative coupling strategy further provides a controllable
accuracy--runtime tradeoff, where the iteration budget can be adjusted to trade accuracy for runtime.
Evaluated on a multi-chiplet system and an advanced packaging system,
DAIST achieves up to $178\times$ speedup over traditional FEM solvers
with mean temperature errors of 0.068\% and 0.323\%, respectively,
while demonstrating cross-topology reuse of block-level models across
structurally distinct package assemblies.
\end{abstract}


\section{Introduction}
\label{sec:intro}

Thermal analysis has become increasingly critical in modern heterogeneous
2.5D and 3D integrated packages.
Unlike traditional monolithic single-die designs, contemporary package
architectures consist of multiple thermally coupled components such as
logic dies, high-bandwidth memory (HBM) stacks, interposers, and interface
materials~\cite{lau2022advpkg, lin2017cowos2}.
Heat generated in one component can propagate across neighboring blocks
through physical contact regions, creating complex thermal interactions
throughout the package.
Accurate evaluation of such multi-block thermal systems is essential for
ensuring reliability, preventing localized thermal hot spots, and enabling
thermal-aware design decisions.

High-fidelity numerical solvers such as the finite-element method (FEM)
and finite-volume method (FVM) are widely used for package-level thermal
analysis in tools such as Ansys IcePak, Cadence Celsius, and
COMSOL~\cite{li2004thermal, chen2017fem, sridhar2010thermal3dic,
sultan2019survey, lin2021thermalfloor, indesignthermal2024}.
While these methods provide accurate solutions to the governing heat
equations, their computational cost grows rapidly with mesh resolution,
material heterogeneity, and system size.
For complex multi-die packages, a single full-package simulation can take
minutes to hours, making repeated thermal evaluation impractical during
iterative design workflows.

With the rapid evolution of machine learning, data-driven and physics-informed
approaches have attracted growing attention for thermal modeling.
Physics-informed neural networks (PINNs)~\cite{raissi2019pinn} incorporate
governing equations into the loss function to improve physical consistency,
but remain tied to fixed design configurations and typically require retraining for
each design.
Operator-learning approaches instead learn mappings between function
spaces and have shown promising performance for 
partial differential equation (PDE) surrogate modeling.
For example, DeepOHeat~\cite{liu2023deepoheat} and DeepOHeat-v1~\cite{yu2025deepoheatv1}
map power and boundary-condition inputs to 3D temperature fields with significant 
speedups over FEM, while PI-ONet~\cite{sha2025pionet} extends this approach
to multilayer chiplet stacks.
Neural operators such as the Fourier Neural Operator (FNO)~\cite{li2021fourier}
and the U-shaped Neural Operator (U-NO)~\cite{rahman2022uno} further improve
data efficiency and scalability for PDE-based surrogate modeling, enabling
significant acceleration of thermal simulations~\cite{huang2025saufno}.

While effective for fixed package configurations, monolithic models
have several limitations for heterogeneous package architectures.
First, the learned model is strongly tied to the entire package topology,
making it difficult to scale trained models across different numbers of dies,
memory stacks, and handle changes in interface regions.
Second, interface physics between neighboring components is implicitly
embedded within the model rather than explicitly enforced.
As a result, temperature continuity 
and heat-flux balance across interfaces cannot be monitored 
or controlled during inference, limiting the model's physical 
interpretability. 

These limitations are particularly problematic for heterogeneous
packages, which naturally consist of multiple physical components
with different materials, geometries, and thermal behaviors.
Instead of learning a single global predictor, it is desirable to
treat each component as an independent thermal block whose behavior
can be modeled separately and flexibly composed into larger package
systems.

Domain decomposition methods (DDM) provide a natural framework for addressing
such coupled multi-domain problems.
In classical DDM, a large computational domain is partitioned into smaller
subdomains, each solved locally while exchanging boundary information with
neighboring subdomains.
Iterative updates of interface boundary conditions ensure that the local
solutions converge to a globally consistent solution ~\cite{yang2014fast3d}.
Recently, several studies have explored combining neural operators with domain
decomposition to accelerate the solution of PDEs ~\cite{mao2024snapddm, wu2025lddm}.
These approaches demonstrate the potential of learned models as local
subdomain solvers within an iterative framework.

However, existing learning-based DDM approaches are primarily developed for
canonical PDE settings, where the global domain is partitioned into
reusable local subproblems or congruent subdomains.
Their main focus is geometry generalization and efficient local surrogate
reuse, rather than modular thermal assembly of heterogeneous blocks.
In contrast, 3D-IC packages are inherently compositional
thermal systems consisting of different blocks.
These blocks may exhibit different materials, internal structures, and
modeling requirements while being strongly coupled through shared
interfaces.
Modeling such systems therefore requires explicit handling of interface
coupling as well as the ability to combine different solver types within a
unified framework.

In this work, we adopt a compositional perspective and formulate 
heterogeneous package thermal analysis as a modular
multi-block system. Each physical component is represented
by an independent block-level thermal predictor, while global
thermal consistency is enforced through iterative exchange of
interface boundary conditions among adjacent blocks. This
formulation allows flexible composition of heterogeneous
blocks when simulating larger package assemblies, and supports
reusable component-level models.

The main contributions of this work are summarized as follows:

\begin{itemize}
  \item We formulate heterogeneous multi-block thermal analysis as a
  compositional thermal modeling framework, where each physical
  component (e.g., dies, HBMs, interposers) is represented by an
  independent block-level thermal predictor connected through physical
  interfaces.
  This formulation enables flexible composition of heterogeneous blocks
  into larger package assemblies, avoiding the tight coupling inherent
  in monolithic AI models that learn the entire package as a single
  predictor.
  \item We propose an interface-iterative thermal solving scheme that
  explicitly exchanges temperature and heat-flux boundary conditions
  between neighboring blocks.
  The iterative formulation enables controllable trade-offs between
  solution accuracy and runtime by adjusting the iteration budget,
  while enforcing temperature continuity and heat-flux balance across
  interfaces.
  \end{itemize}
  
\section{Methodology}
\label{sec:method}

\subsection{Thermal Problem Formulation}

This work considers steady-state heat conduction in heterogeneous multi-block systems.
Let $\Omega$ denote the global thermal domain composed of multiple
physically connected blocks.
The temperature distribution $T(\mathbf{x})$ satisfies the heat
conduction equation,

\begin{equation}
\nabla \cdot (k \nabla T) + Q = 0,
\end{equation}

where $k$ denotes the thermal conductivity and $Q$ represents the
internal heat generation. The global domain $\Omega$ is partitioned into $N$
non-overlapping thermal blocks, as described in Eq.~\ref{eq:block_partition}.

\begin{equation}
\Omega = \bigcup_{b=1}^{N} \Omega_b, \qquad
\Omega_i \cap \Omega_j = \emptyset \quad (i \neq j)
\label{eq:block_partition}
\end{equation}

Each block in Eq.~\ref{eq:block_partition} corresponds to a physical component in the
package, such as a logic die, HBM stack, interposer, etc.
Therefore, the decomposition follows the physical assembly
structure of heterogeneous packages rather than
using any artificial computational partitions.

Standard thermal boundary conditions are applied to external
boundaries, including Dirichlet (fixed temperature), Neumann
(prescribed heat flux), and Robin (convective cooling) conditions. Blocks interact through shared physical interfaces.
For each interface $\Gamma_{ij}$ between neighboring blocks
$\Omega_i$ and $\Omega_j$, the physical solution must satisfy

\begin{itemize}
\item temperature continuity: $T_i = T_j$ on $\Gamma_{ij}$,
\item heat-flux balance: $k_i \nabla T_i \cdot n_i =
      -k_j \nabla T_j \cdot n_j$ on $\Gamma_{ij}$,
\end{itemize}

which ensure global thermal consistency across block boundaries.

\subsection{Block Decomposition and Interface Coupling}

To enable modular thermal solving, the global domain is decomposed into
individual blocks that are solved locally.
Each block receives boundary information from its neighboring blocks through
shared interfaces. For each interface $\Gamma_{ij}$ between two blocks,
the temperature and heat flux are exchanged to serve as updated boundary conditions for the next iteration. At iteration $k$, each block receives interface boundary conditions
computed from its neighbors during iteration $k-1$. To be specific, in the proposed coupling scheme, temperature at the interface is passed from one block
to its neighbor as a Dirichlet boundary condition, while the receiving
block returns the corresponding heat flux at the same interface. 

\subsection{Interface-Iterative Thermal Solver}
\label{sec:iterative_solver}

\begin{figure}[t]
\centering
\resizebox{\columnwidth}{!}{%
\begin{tikzpicture}[
    >=Stealth,
    block/.style={draw, fill=blue!12, rounded corners=2pt,
                  minimum width=1.4cm, minimum height=0.55cm,
                  font=\scriptsize, align=center},
    volblock/.style={draw, fill=teal!15, rounded corners=2pt,
                     minimum width=1.5cm, minimum height=0.55cm,
                     font=\scriptsize, align=center},
    check/.style={draw, fill=orange!18, rounded corners=2pt,
                  minimum width=1.6cm, minimum height=0.6cm,
                  font=\scriptsize, align=center},
    output/.style={draw, fill=gray!15, rounded corners=2pt,
                   minimum width=1.0cm, minimum height=0.6cm,
                   font=\scriptsize, align=center},
    arr/.style={->, thin, gray!70},
    xline/.style={gray!50, thin},
]
 
% ── Row y-coordinates ──
\def\rowA{2.0}
\def\rowB{1.0}
\def\rowC{0.0}
 
% ── Block labels ──
\node[font=\scriptsize, anchor=east] at (-0.15, \rowA) {$\Omega_1$};
\node[font=\scriptsize, anchor=east] at (-0.15, \rowB) {$\Omega_2$};
\node[font=\scriptsize, anchor=east] at (-0.15, \rowC) {$\Omega_3$};
 
% ── Phase 1 container ──
\draw[dashed, rounded corners=6pt, gray]
     (0.05, -0.5) rectangle (3.75, 2.6);
\node[font=\footnotesize, anchor=south] at (1.9, 2.65)
     {Interface iteration};
\node[font=\scriptsize, gray, anchor=north] at (1.9, -0.55)
     {Predict surface $T$, $q$};
 
% ── Column sub-labels ──
\node[font=\scriptsize, gray] at (1.0, 2.45) {Iter $k$};
\node[font=\scriptsize, gray] at (2.9, 2.45) {Iter $k{+}1$};
 
% ── Iter k boxes ──
\node[block] (a1) at (1.0, \rowA) {AI model};
\node[block] (a2) at (1.0, \rowB) {AI model};
\node[block] (a3) at (1.0, \rowC) {AI model};
 
% ── Iter k+1 boxes ──
\node[block] (b1) at (2.9, \rowA) {AI model};
\node[block] (b2) at (2.9, \rowB) {AI model};
\node[block] (b3) at (2.9, \rowC) {AI model};
 
% ── BC exchange crossing lines ──
\draw[xline] (a1.east) -- (b2.west);
\draw[xline] (a2.east) -- (b1.west);
\draw[xline] (a2.east) -- (b3.west);
\draw[xline] (a3.east) -- (b2.west);
\draw[xline, dashed] (a1.east) -- (b1.west);
\draw[xline, dashed] (a3.east) -- (b3.west);
 
% ── Omega label ──
\node[font=\scriptsize] at (1.95, 0.75) {$\omega$};
 
% ── Dots ──
\node[font=\normalsize] at (4.15, \rowB) {$\cdots$};
 
% ── Convergence check (shifted right for arrow clearance) ──
\node[check] (conv) at (5.5, \rowB) {$\delta^{(k)} < \tau\,$?};
 
% ── Arrow: dots → convergence ──
\draw[arr] (4.4, \rowB) -- (conv.west);
 
% ── No loop: up → left → down into Phase 1 ──
\draw[arr] (conv.north) -- ++(0, 2.0)
           -- (1.9, 3.3) -- (1.9, 2.55);
\node[font=\scriptsize, anchor=south] at (3.7, 3.3) {No};
 
% ── Phase 2 container ──
\draw[dashed, rounded corners=6pt, gray]
     (6.8, -0.5) rectangle (8.6, 2.6);
\node[font=\footnotesize, anchor=south] at (7.7, 2.65)
     {Field reconstruction};
\node[font=\scriptsize, gray, anchor=north] at (7.7, -0.55)
     {Predict full 3-layer $T$};
 
% ── Phase 2 boxes ──
\node[volblock] (v1) at (7.7, \rowA) {AI model};
\node[volblock] (v2) at (7.7, \rowB) {AI model};
\node[volblock] (v3) at (7.7, \rowC) {AI model};
 
% ── Yes arrow: conv → Phase 2 ──
\draw[arr] (conv.east) -- (6.8, \rowB)
           node[font=\scriptsize, above, midway] {Yes};
 
% ── Output ──
\node[output] (out) at (9.4, \rowB) {$T(\mathbf{x})$};
\draw[arr] (v1.east) -- (out.north west);
\draw[arr] (v2.east) -- (out.west);
\draw[arr] (v3.east) -- (out.south west);
 
\end{tikzpicture}
}
\caption{Overview of the DAIST two-phase algorithm.
\textbf{Interface iteration} (left): each block's AI model predicts
surface temperature and heat flux, and interface quantities are exchanged
between neighboring blocks with relaxation parameter~$\omega$. The
process repeats until convergence ($\delta^{(k)} < \tau$).
\textbf{Field reconstruction} (right): a single forward pass per
block produces the full three-layer temperature field, which is
assembled into the global solution field ~$T(\mathbf{x})$.}
\label{fig:algorithm}
\end{figure}
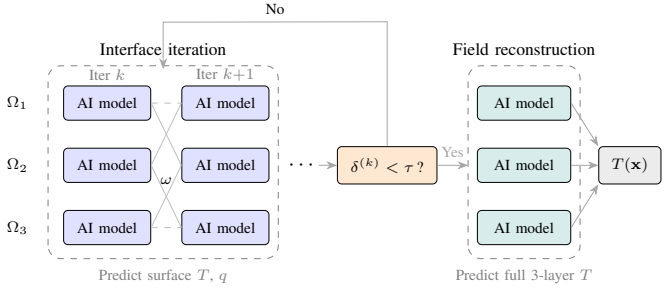

The global thermal solution is obtained through a two-phase
iterative procedure illustrated in Figure~\ref{fig:algorithm}.

\paragraph{Phase 1 (interface iteration)}
At each iteration~$k$, every block solves a \emph{boundary-level}
subproblem with boundary conditions obtained from the last-iteration interface quantities or initial-step conditions. Given the interface conditions from neighboring blocks, the local solver returns updated interface temperature and
heat flux at shared surfaces.
These interface quantities are exchanged between neighboring blocks
and smoothed via relaxation:
\begin{equation}
  u^{(k)} = u^{(k-1)}
           + \omega_u \!\left( u^{(k-1)}_{\mathrm{neighbor}}
                             - u^{(k-1)} \right),
  \label{eq:relaxation}
\end{equation}
where $\omega_u \in (0,1]$ is a variable-specific relaxation parameter
whose role is formalized in Section~\ref{subsec:solver_interpretation}.
The interface iteration repeats until the mean temperature change
across all interfaces satisfies
\begin{equation}
  \delta^{(k)} < \tau,
  \label{eq:convergence_criterion}
\end{equation}
where $\tau$ is a user-specified tolerance. In this phase, only surface quantities are predicted, so each local solve is lightweight.

\paragraph{Phase 2 (field reconstruction)}
Once interface convergence is achieved, a single \emph{volume-level}
forward pass is executed: each block solver takes the converged
interface boundary conditions and produces the full three-layer
temperature field (substrate, device, interconnect) within that
block. The per-block fields are then assembled into the global
temperature solution.

The two-phase design keeps the iterative loop inexpensive by
restricting repeated evaluation to interface quantities, while the
one-shot field reconstruction amortizes the cost of generating the
full volumetric output.

% ============================================================
%  TikZ diagram (reproduces the SVG overview for the paper)
% ============================================================
% Requires: \usepackage{tikz}
%           \usetikzlibrary{positioning, fit, arrows.meta, 
%                           decorations.pathreplacing, calc}

\subsection{Local Block Solvers}
\label{subsec:local_solvers}

Within the proposed framework, each block is solved independently
using a local solver, which is an AI-based surrogate model in this work. 

The local solver takes the following as input: 

\begin{itemize}
\item power distribution within the block
\item interface boundary conditions from neighboring blocks
\item block-specific geometry and material parameters
\end{itemize}

The solver outputs updated interface quantities (temperature or heat flux)
after solving the local thermal problem.

In this work, AI-based local solvers are instantiated using a neural
operator architecture. Specifically, we employ the U-NO~\cite{rahman2022uno} to approximate the mapping between
block inputs and interface responses. The proposed framework is not
tied to a specific model architecture, and other surrogate models
(e.g., CNN-based models or physics-informed networks) can also serve
as local block solvers within the same interface-iterative procedure.

\subsection{Block Representation}

To reduce modeling complexity while preserving dominant thermal
behavior, each block is represented using a simplified layered
structure consisting of a substrate layer, a device layer, and an interconnect layer. 
The interconnect layer aggregates multiple back-end-of-line (BEOL) metal and via
layers into a single effective thermal layer.
This representation captures the dominant vertical thermal resistance
while maintaining manageable input dimensionality for the learned
solver.

\subsection{Solver Interpretation}
\label{subsec:solver_interpretation}
The proposed framework can be interpreted as 
a block-iterative linear solver for the discretized thermal system. After spatial discretization, the global thermal problem becomes
\begin{equation}
  A\,x = b
  \label{eq:global}
\end{equation}
where $x \in \mathbb{R}^{n}$ collects the temperature unknowns across 
all blocks. Partitioning $x$ by block
$x = [x_1^\top, \ldots, x_N^\top]^\top$
induces a block decomposition
\begin{equation}
  A = D + R, \qquad
  D = \operatorname{blkdiag}(A_{11}, \ldots, A_{NN})
  \label{eq:split}
\end{equation}
where each diagonal block $A_{bb}$ captures internal thermal
coupling within block~$\Omega_b$, and the off-diagonal remainder
$R$ is nonzero only at shared interface nodes.
Isolating block~$b$ gives
$A_{bb}\,x_b = b_b - \textstyle\sum_{i \neq b} A_{bi}\,x_i$,
showing that each block can be solved independently once its
neighbor interface contributions are known.
 
The iterative update in~\eqref{eq:relaxation} corresponds to a
\emph{relaxed block Jacobi} iteration~\cite{saad2003iterative}:
\begin{equation}
  x^{(k)} = (1-\omega)\,x^{(k-1)}
           + \omega\,D^{-1}\!\left(b - R\,x^{(k-1)}\right)
  \label{eq:jacobi}
\end{equation}
Because $D$ is block-diagonal, the inversion $D^{-1}$ decouples into
$N$ independent local solves that can execute in parallel. 

With exact numerical solvers for each block, this recovers a classical
non-overlapping domain decomposition iteration~\cite{yang2014fast3d}.
When learned surrogates replace exact solves, the framework
becomes an \emph{inexact} block Jacobi iteration. Convergence is
preserved provided the per-block surrogate error remains small
relative to the interface residual~\cite{saad2003iterative}. 
% Global consistency can be achieved as long as the iterative treatment described in Fig.~\ref{fig:algorithm} is followed.
\section{Experimental Results}
\label{sec:results}

% ============================================================
\subsection{Experimental Setup}
\label{subsec:experiment_setup}

\begin{figure}[t]
  \centering
  \includegraphics[width=\columnwidth]{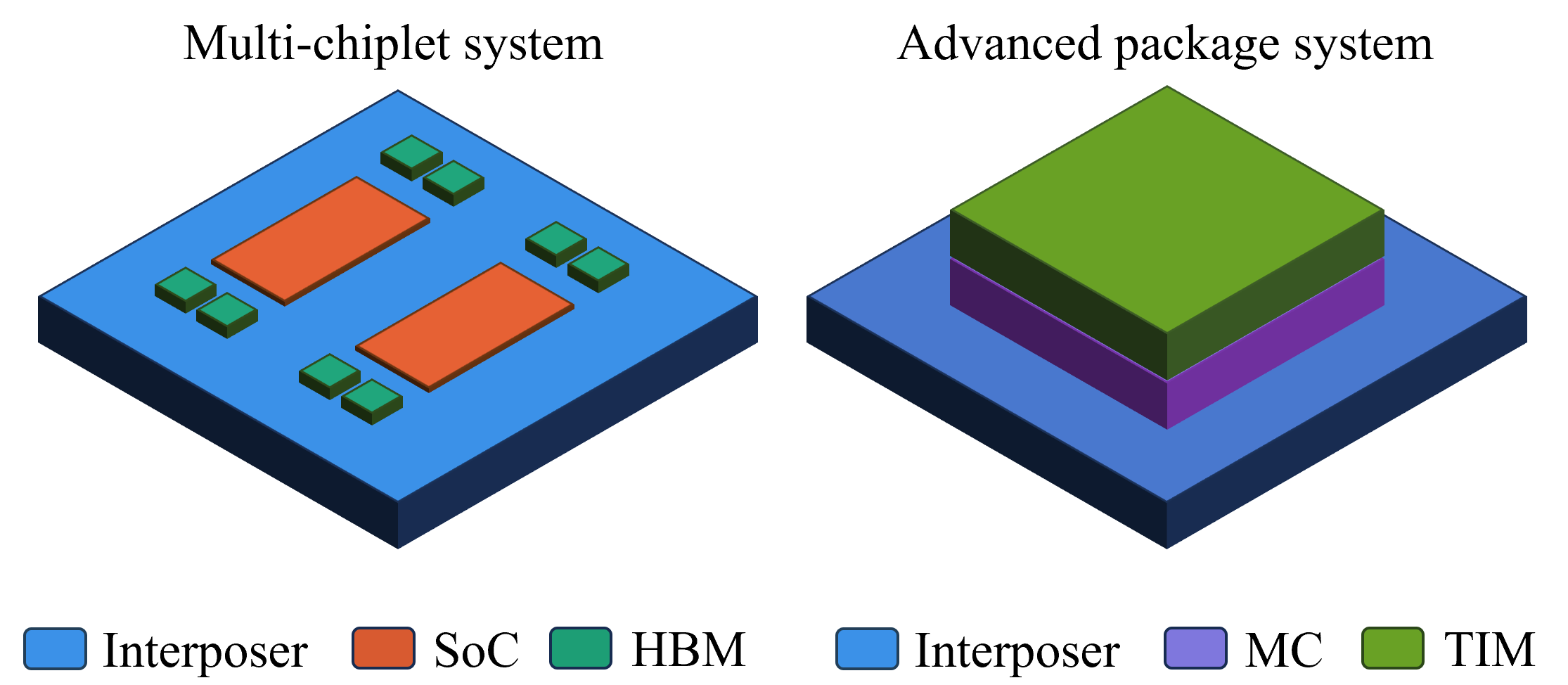}
  \caption{Representative stackup schematics (not to scale) for the multi-chiplet assembly
  (Interposer--SoC--HBM, left) and the advanced packaging assembly (Interposer--MC--TIM, right):
  planar SoC and HBM placements on the shared interposer versus a vertical stack of molding
  compound (MC) and thermal interface material (TIM), consistent with the block parameters
  in Table~\ref{tab:geometric_parameters}.}
  \label{fig:stackup}
\end{figure}

\noindent\textbf{Case Studies and Configurations.}
We evaluate DAIST on two representative 2.5D/3D-IC assemblies:
the multi-chiplet system (Interposer--SoC--HBM) and the advanced packaging
system (Interposer--MC--TIM), as depicted in Figure~\ref{fig:stackup}.
In the multi-chiplet system, multiple SoC and HBM stacks are mounted on a shared interposer
in a flip-chip assembly, with SoC and HBM dies attached face-down to the face-up interposer.
To demonstrate composability, SoC instance counts are sampled uniformly from $\{1,2,3\}$
and HBM counts are sampled uniformly from $\{1,\ldots,10\}$. This sampling strategy produces assemblies with substantially diverse thermal footprints and interface coupling complexity.
In the advanced packaging system, the molding compound (MC) is positioned face-down
and covers the region of interest on the interposer, while the thermal interface material (TIM)
is placed above the MC.
Each block uses the three-layer abstraction from Section~\ref{sec:method}
(substrate, device, interconnect).
Layer thicknesses, conductivities $k$, and volumetric heat generation
follow Table~\ref{tab:geometric_parameters}.
Each layer follows a $50\times50$ in-plane discretization.

\begin{table*}[t]
  \centering
  \caption{Block geometry and material parameters for Multi-chiplet System and Advanced Packaging System.}
  \label{tab:geometric_parameters}
  \setlength{\tabcolsep}{4pt}
  \small
  \begin{tabular}{@{}lllcccc@{}}
  \toprule
  \textbf{Block} & \textbf{Orientation} & \textbf{No.\ components} & \textbf{Footprint}$^\ddagger$
    & \textbf{Thickness ($\mu$m)$^\dagger$} & \textbf{$k$ (W/m$\cdot$K)$^\dagger$} & \textbf{Power (W)$^\dagger$$^\ast$} \\
  \midrule
  \multicolumn{7}{@{}l}{\emph{Case I: Multi-chiplet system (Interposer--SoC--HBM)}} \\
  \midrule
  Interposer & face up   & $1$        & $5\times 5$ mm      & $(100,\ 1,\ 1)$  & $(100,\ 2.175,\ 0.3)$ & $(0,\ 0,\ 0)$ \\
  SoC        & face down & $[1,\ 3]$  & $2\times 1$ mm      & $(10,\ 1,\ 1)$   & $(100,\ 2.175,\ 0.3)$ & $(0,\ [0.05, 0.1],\ 0)$ \\
  HBM        & face down & $[1,\ 10]$ & $0.5\times 0.4$ mm  & $(20,\ 1,\ 1)$   & $(100,\ 2.175,\ 0.3)$ & $(0,\ [0.01, 0.03],\ 0)$ \\
  \midrule
  \multicolumn{7}{@{}l}{\emph{Case II: Advanced packaging system (Interposer--MC--TIM)}} \\
  \midrule
  Interposer & face up   & $1$ & $5\times 5$ mm & $(100,\ 1,\ 1)$ & $(100,\ 2.175,\ 0.3)$ & $(0,\ 0,\ 0)$ \\
  MC         & face down & $1$ & $3\times 3$ mm & $(100,\ 1,\ 1)$ & $(100,\ 2.175,\ 0.3)$ & $(0,\ [0.008, 0.08],\ 0)$ \\
  TIM        & face up   & $1$ & $3\times 3$ mm & $(100,\ 1,\ 1)$ & $(67,\ 67,\ 67)$      & $(0,\ 0,\ 0)$ \\
  \bottomrule
  \end{tabular}\\[0.3em]
  \footnotesize
  $^\dagger$Values listed as (substrate layer, device layer, interconnect layer) triples.\\
  $^\ast$Power is listed as a range of values in [min, max] format.\\
  $^\ddagger$Footprint is in-plane ($x\times y$).
\end{table*}

\smallskip\noindent\textbf{Boundary Conditions.}
Adiabatic (Neumann) conditions with zero normal heat flux are imposed on all exterior lateral faces.
For both assemblies, top and bottom surfaces have a heat transfer coefficient (HTC) of $h=100$~W/(m$^2\cdot$K),
with the exception of the top surface of the interposer, which is adiabatic.
For all surfaces with HTC, we use a fixed ambient temperature $T_\infty = 298.15$~K.
Internal block-to-block interfaces follow the iterative Dirichlet/Neumann exchange
described in Section~\ref{sec:method} without separate contact resistances.

\smallskip\noindent\textbf{Data Generation.}
For each block type (interposer, SoC, HBM, MC, TIM), a separate supervised
dataset is generated with geometric parameters consistent with Table~\ref{tab:geometric_parameters}.
In the multi-chiplet system, power is generated exclusively in the SoC and HBM blocks, with
the interposer carrying no internal heat sources.
Power maps are sampled from Gaussian random fields (GRFs) with length-scale parameter $0.3$.
In the advanced packaging system, the interposer and TIM have no internal heat sources, and the MC block
contains 1--4 embedded power regions at random locations. Each power region has a lateral dimension of 1\,mm $\times$ 1\,mm, with
power profiles and interface boundary conditions sampled from the same GRF as described above.
Each block dataset contains 5{,}000 total samples.
For each sample, ground-truth fields are obtained from the Celsius Thermal Solver (abbreviated as Celsius). Footprints, layer thicknesses, conductivities, and external boundary conditions can all be
treated as input variables for training. For simplicity, these variables are fixed in this work
because variations in power, internal interface boundary conditions, and topology are the primary focus.

\smallskip\noindent\textbf{Training.}
For each block dataset, a ratio of 8:1:1 is used for splitting into training, validation, and test sets. This yields 4{,}000 training, 500 validation, and 500 test samples per block type.
The interface relaxation parameters are set to $\omega_T = 0.5$ for temperature
and $\omega_q = 0.5$ for heat flux, which are selected via grid search on the validation set.
Each model is trained for 1,000 epochs on 4 NVIDIA A100 GPUs, taking approximately 3 hours per block type. Training all five block types (interposer, SoC, HBM, MC, TIM) requires 15 hours in total, but these models are subsequently reused across any topology without retraining. By contrast, a monolithic predictor requires a full training run for each new package configuration, as its learned mapping is tied to a specific die count and arrangement.

\smallskip\noindent\textbf{Compared Methods.}
DAIST results were compared against the Celsius Thermal Solver and a monolithic
neural predictor (ML baseline).
Existing multi-block ML baselines for thermal simulation are limited. For example, single-block methods such as DeepOHeat-v1~\cite{yu2025deepoheatv1} and
PI-ONet~\cite{sha2025pionet} address fixed single-domain prediction and
cannot extend to variable-topology multi-block assemblies without fundamental
architectural changes and full retraining.
Therefore, a monolithic U-NO predictor is adopted in this work as the ML baseline for multi-block
thermal analysis. It is the most direct neural alternative to DAIST for full-package
prediction, and it uses an identical architecture and hyperparameters to DAIST's
per-block models.
A detailed description and analysis of this baseline is provided in
Section~\ref{subsec:generalization}.

% ============================================================
\subsection{Multi-Chiplet System: Accuracy and Runtime}
\label{subsec:multi-chiplet-system}

\begin{figure*}[t]
  \centering
  \includegraphics[width=\textwidth]{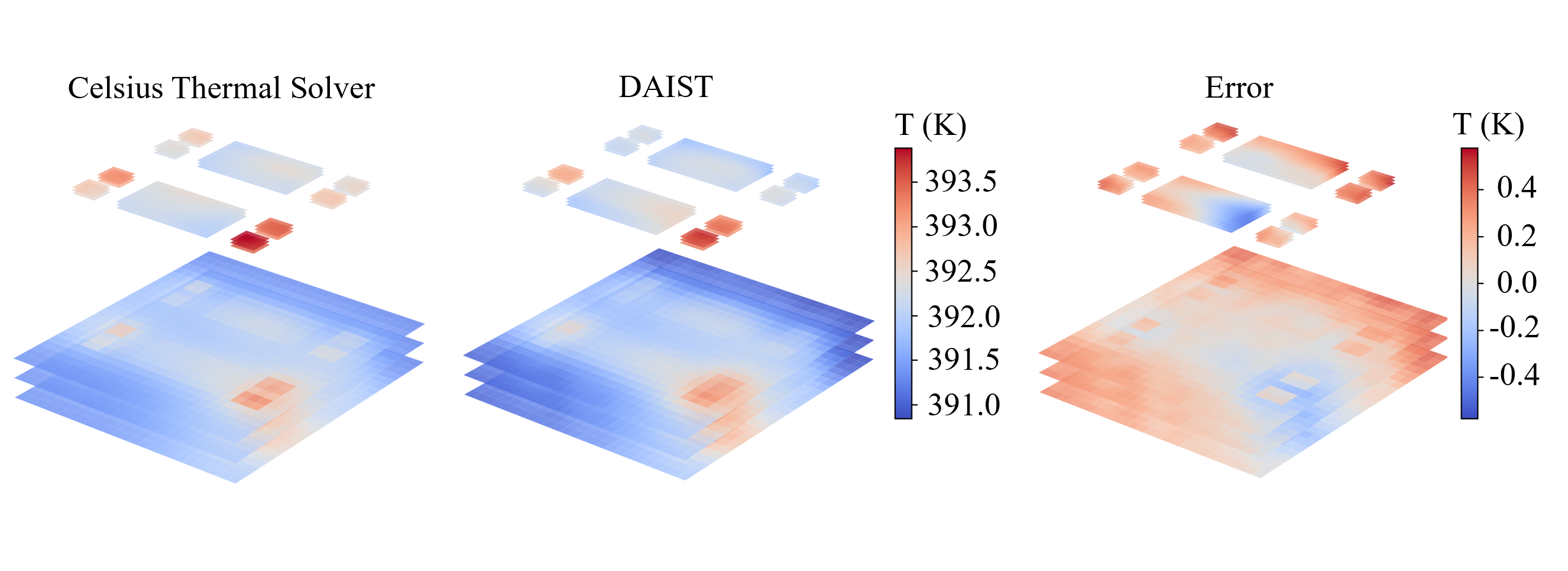}
  \caption{Three-dimensional thermal visualization of the multi-chiplet assembly (Interposer--SoC--HBM).
    Left: Celsius Thermal Solver reference. Center: DAIST prediction. Right: pointwise absolute error.}
  \label{fig:multi_chiplet_3d}
\end{figure*}

Figure~\ref{fig:multi_chiplet_3d} shows a three-dimensional temperature visualization
of the multi-chiplet (Interposer--SoC--HBM) assembly with an example stackup of
2~SoC and 8~HBM blocks.
DAIST's prediction closely matches the Celsius reference across all blocks and layers.
The interposer exhibits a thermal gradient from approximately 391\,K at the periphery
to 393.5\,K beneath the high-power SoC footprints, with both methods capturing the
same hotspot locations and magnitudes.
The pointwise error field is within $\pm 0.5$\,K, with over 70\% of pointwise errors
below $0.2$\,K. The largest errors are localized at die-interposer interface regions
where thermal gradients are steepest.

Table~\ref{tab:multi_chiplet_summary} summarizes DAIST's accuracy and runtime
on 100 held-out test cases from the multi-chiplet system, using the Celsius reference as the ground truth. DAIST achieves an MAE of $0.253$\,K, RMSE of $0.341
$\,K, MAPE of $6.79\times10^{-4}$, and an average
per-case maximum error of $1.135$\,K, while delivering a $178\times$ speedup
over the full-package Celsius solve. A comprehensive comparison against the
monolithic neural predictor, including both in-distribution and
out-of-distribution evaluation, is presented in
Section~\ref{subsec:generalization}.

\begin{table}[t]
  \centering
  \caption{Accuracy and runtime comparison on the multi-chiplet system.}
  \label{tab:multi_chiplet_summary}
  \small
  \begin{tabular}{@{}lcccc@{}}
    \toprule
    \textbf{Method} & \textbf{MAE (K)} & \textbf{Max (K)} & \textbf{Runtime (s)} & \textbf{Speedup} \\
    \midrule
    Celsius reference  & ---   & ---   & $146$  & $1\times$    \\
    DAIST          & $\mathbf{0.253}$ & $\mathbf{1.135}$ & $0.82$  & $\mathbf{178\times}$  \\
    \bottomrule
  \end{tabular}
\end{table}

\smallskip\noindent\textbf{Speedup.}
A full-package Celsius solve takes $146$\,s per case.
DAIST's per-iteration boundary update costs approximately $0.117$\,s on one NVIDIA A100 GPU.
With a predefined stopping criterion of 0.1\% relative change between consecutive iterations,
DAIST requires on average $5.2$ boundary iterations ($\approx 0.61$\,s) plus one final
full-field reconstruction ($0.091$\,s), for an end-to-end inference time of $0.82$\,s per
case. This cost includes data-transfer and convergence-check overhead, and it is a speedup of $\mathbf{178\times}$ over Celsius.
Because each block model is solved independently, all inferences of the same block type
(interposer, SoC, HBM) within a single package are batched into one GPU call, making
per-package latency largely independent of block count.

\smallskip\noindent\textbf{Accuracy--Runtime Tradeoff via Iteration Control.}
As discussed above, DAIST converges in 5.2 iterations on average under the default stopping criterion. However, users can manually control the tradeoff between accuracy and runtime. Table~\ref{tab:multi_chiplet_iter_mae} shows the mean MAE on the 100-case test set
when interface coupling is stopped after a fixed number of iterations (no early stopping). MAE decreases sharply through the first four iterations and saturates by iteration six. Table~\ref{tab:multi_chiplet_iter_mae} extends to seven iterations to confirm this. Notably, even at the highest accuracy (seven iterations), the wall time remains well below the 146\,s Celsius reference. This allows users to control the iteration budget to balance runtime and precision requirements.

\begin{table}[t]
  \centering
  \caption{Mean MAE and wall time at fixed iteration counts for multi-chiplet test set.}
  \label{tab:multi_chiplet_iter_mae}
  \small
  \begin{tabular}{@{}ccc@{}}
    \toprule
    \textbf{Iterations} & \textbf{Mean MAE (K)} & \textbf{Wall time$^*$} \\
    \midrule
    $1$ & $4.228$ & $0.328$ \\
    $2$ & $3.398$ & $0.445$ \\
    $3$ & $1.769$ & $0.562$ \\
    $4$ & $0.757$ & $0.679$ \\
    $5$ & $0.290$ & $0.796$ \\
    $6$ & $0.168$ & $0.913$ \\
    $7$ & $0.159$ & $1.030$ \\
    \midrule
    Celsius reference & --- & $146$ \\
    \bottomrule
  \end{tabular}\\[0.3em]
  \footnotesize $^*$Includes field construction and other overhead time.
\end{table}

\begin{figure}[t]
  \centering
  \includegraphics[width=\columnwidth]{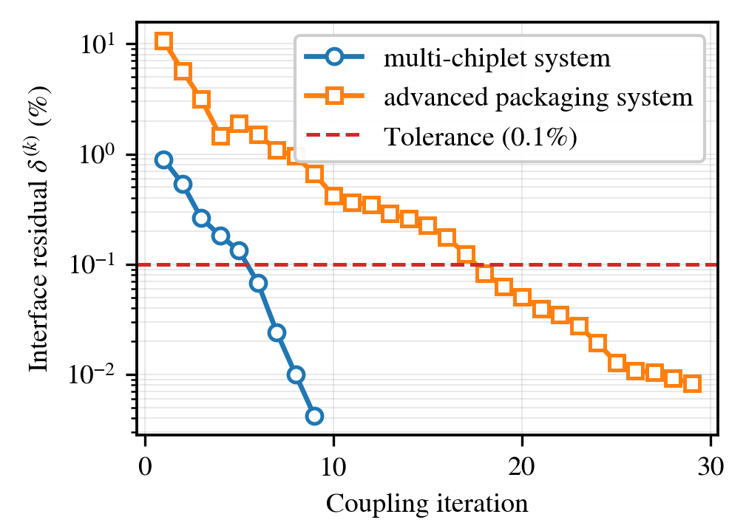}
  \caption{Interface residual $\delta^{(k)}$ vs.\ coupling iteration
    for both case studies. The dashed line marks the convergence
    tolerance $\tau = 0.1\%$.}
  \label{fig:convergence}
\end{figure}

Figure~\ref{fig:convergence} shows the convergence history of the
interface residual $\delta^{(k)}$ for both case studies.
Note that $\delta^{(k)}$ measures the change between consecutive iterations
rather than the error against the ground-truth solution, so it decreases
more steadily than MAE.
The multi-chiplet system reaches the $0.1\%$ tolerance in approximately 5 iterations,
while the advanced packaging system requires approximately 15 iterations.
This is due to the difference in the coupling topology.
In the multi-chiplet case, all dies share direct interfaces with the interposer,
whereas the sequential chain (interposer $\to$ MC $\to$ TIM) requires multiple
passes for thermal information to propagate end-to-end.
The advanced packaging system is described in detail in Section~\ref{subsec:cross_topology}.
Both cases exhibit stable and monotonic convergence, with appropriate
tuning of the relaxation parameter $\omega$.

% ============================================================
\subsection{Comparison with Monolithic Neural Predictor}
\label{subsec:generalization}

DAIST is evaluated against the monolithic neural predictor, the most direct
ML baseline for full-package thermal prediction, on the multi-chiplet system. This comparison
focuses on whether compositional decomposition preserves in-distribution
accuracy while enabling generalization to unseen topologies.

\smallskip\noindent\textbf{Setup.}
A fixed-topology dataset of 5{,}000 cases is constructed with the 2\,SoC\,+\,8\,HBM
configuration, and split by 8:1:1 into training, validation, and test sets.
Locations and power maps of SoC and HBM blocks are randomized across all samples.
Both DAIST and the monolithic predictor are trained using this dataset.
This DAIST instance is trained under the same fixed-topology constraint as the
monolithic predictor. It is distinct from the general
DAIST evaluated in Section~\ref{subsec:multi-chiplet-system}, which is trained across
variable topology counts. Accuracy metrics between the two subsections are therefore
not directly comparable.

The monolithic predictor uses an identical U-NO architecture and hyperparameters to
each per-block model in DAIST.
Its inputs are the SoC power maps, HBM power maps, SoC footprint masks, HBM footprint
masks, and external boundary conditions. Its output is the interposer top-surface
temperature field.
SoC and HBM masks are kept separate because they differ in height, power range, and
thermal pathways.
Notably, the monolithic predictor receives strictly richer input than any individual
DAIST block model, including full power maps and spatial masks for all dies. This gives the monolithic predictor an
information advantage per forward pass.
DAIST's per-block models each receive only their local power map and interface boundary
conditions, with no global topology information.
This makes the comparison conservative for DAIST. If the monolithic predictor still
fails on unseen topologies despite having access to more information, the failure is
structural and cannot be attributed to any information asymmetry favoring DAIST.

\smallskip\noindent\textbf{Evaluation.}
Both methods are evaluated on two test sets of 100 cases each:
\begin{itemize}
  \item \textbf{Fixed-topology}: held-out cases from the same 2\,SoC\,+\,8\,HBM
    distribution (in-distribution).
  \item \textbf{Variable-topology}: cases drawn from the general multi-chiplet
    distribution in Section~\ref{subsec:experiment_setup}, where SoC count is
    sampled uniformly from $\{1,2,3\}$ and HBM count from $\{1,\ldots,10\}$.
    This test set is out-of-distribution for \emph{both} methods, as neither
    was retrained on variable topologies.
\end{itemize}
Although DAIST predicts temperature fields for all blocks, we report only the
interposer top-surface temperature for fair comparison, as this is the only
quantity both methods produce.
Results are reported in Table~\ref{tab:benchmark_daist_monolithic}.

\begin{table*}[t]
  \centering
  \caption{Accuracy and runtime comparison of DAIST and the monolithic neural predictor
    on fixed-topology and variable-topology test sets.\textsuperscript{$\dagger$}}
  \label{tab:benchmark_daist_monolithic}
  \setlength{\tabcolsep}{5pt}
  \small
  \begin{tabular}{@{}llccccc@{}}
    \toprule
    \textbf{Method} & \textbf{Test Set} & \textbf{Runtime (s)} &
      \textbf{MAE (K)} & \textbf{Max (K)} & \textbf{MAPE} & \textbf{RMSE (K)} \\
    \midrule
    \multirow{2}{*}{DAIST}       & Fixed-topology    & \multirow{2}{*}{$0.82$} &
      $\mathbf{0.282}$ & $1.230$ & $7.01\times10^{-4}$ & $0.356$ \\
           & Variable-topology &  &
      $\mathbf{0.351}$ & $\mathbf{1.233}$ & $9.10\times10^{-4}$ & $0.415$ \\
    \midrule
    \multirow{2}{*}{Monolithic}  & Fixed-topology    & \multirow{2}{*}{$0.12$} &
      $0.433$ & $\mathbf{0.937}$ & $1.07\times10^{-3}$ & $0.474$ \\
      & Variable-topology &  &
      $26.264$ & $26.944$ & $6.83\times10^{-2}$ & $26.269$ \\
    \bottomrule
  \end{tabular}
  \vspace{2pt}\\
  \parbox{\textwidth}{\footnotesize
  \textsuperscript{$\dagger$}MAE: mean absolute error; Max: mean of per-case maximum absolute
  pointwise error; MAPE: mean absolute percentage error relative to ground truth temperature;
  Runtime: wall-clock inference time per case.}
\end{table*}

\begin{figure*}[t]
  \centering
  \includegraphics[width=\textwidth]{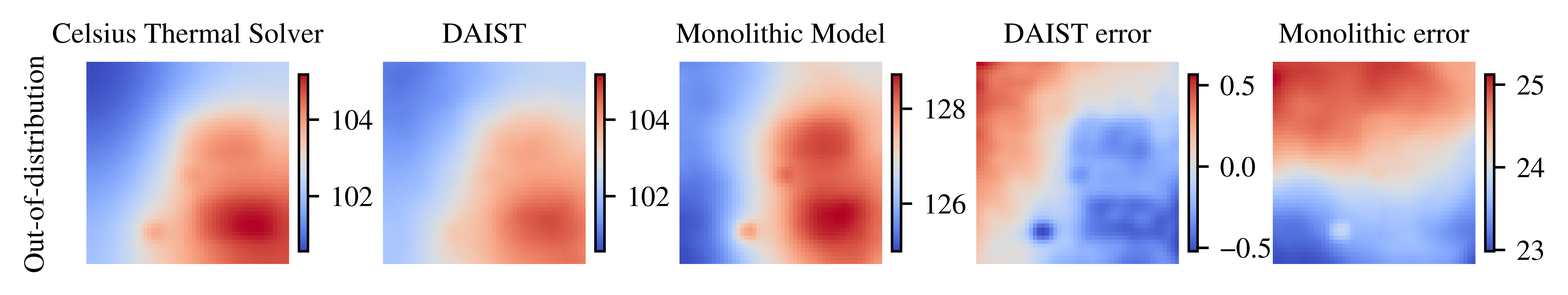}
  \caption{Out-of-distribution comparison on a variable-topology case.
    Left to right: Celsius Thermal Solver reference, DAIST prediction, monolithic prediction,
    DAIST error, monolithic error.}
  \label{fig:ood_comparison}
\end{figure*}

\smallskip\noindent\textbf{In-Distribution Performance.}
On the fixed-topology test set, DAIST achieves lower MAE ($0.282$\,K vs.\ $0.433$\,K)
than the monolithic predictor.
The monolithic predictor is faster at inference ($0.12$\,s vs.\ $0.82$\,s) since it
requires no iterative interface coupling.
Both methods are accurate enough for practical use on the trained configuration,
confirming that iterative decomposition does not sacrifice in-distribution accuracy.
The monolithic predictor's lower latency stems from the absence of iterative coupling
overhead. Users with fixed topology and in-distribution operating conditions
may prefer this increased speed in exchange for less flexibility.
Conversely, DAIST’s iterative overhead is the necessary cost to achieve topology-agnostic generalization.

\smallskip\noindent\textbf{Out-of-Distribution Generalization.}
The critical distinction emerges on the variable-topology test set.
DAIST's MAE increases modestly from $0.282$\,K to $0.351$\,K (24\% relative degradation).
The monolithic predictor degrades catastrophically, with MAE rising from $0.433$\,K to
$26.264$\,K, a $60\times$ increase showing its inability in generalization tasks.

This performance distinction arises from a fundamental structural difference.
The monolithic predictor maps the power distributions of exactly 2~SoC and 8~HBM dies
to a full-package temperature field.
When the die count changes, it receives inputs of a different structure than it was
trained on and fails to generalize.
DAIST decomposes the problem along physical block boundaries, and each block model receives
only its local power map and the interface boundary conditions from neighboring blocks.
Neither the SoC nor the HBM model has information about how many sibling dies exist
on the interposer, but rather they just respond to the flux or temperature at interfaces.
The interposer model likewise receives flux contributions from whatever blocks contact it,
regardless of their count or arrangement.
This localized physical representation enables DAIST to generalize to topologies never encountered during training.

DAIST's modest OOD degradation ($0.282$\,K to $0.351$\,K) shows extrapolation ability at
the individual block-level. In unseen topologies, the interposer receives interface
flux patterns outside the training distribution. 
Nevertheless, the interposer model generalizes well because the underlying
thermal response, governed by the heat equation, maintains a consistent input–-output structure. Locally, higher heat flux leads to proportionally higher temperatures,
and globally, the aggregated heat flux remains linearly related to the
mean temperature rise through an effective thermal resistance.
Since these relationships are independent of die count or placement, the learned operator can extrapolate to unseen
flux patterns without relying on topology-specific features.

Figure~\ref{fig:ood_comparison} visualizes this contrast on a representative
variable-topology case.
The Celsius reference and DAIST prediction show close agreement in both temperature
magnitude and hotspot location, with DAIST errors within $\pm 0.5$\,K across most
of the field.
The monolithic predictor captures the general spatial pattern of hotspot locations,
which is expected since the correlation between local power input and
temperature rise is straightforward to learn. However, it overestimates the absolute
temperature by approximately 24\,K, indicating that the model fails to correctly
resolve the effective thermal resistance of the unseen topology. This confirms
that the monolithic predictor's failure is not in spatial feature extraction but
in its inability to generalize the global thermal balance to configurations
outside the training distribution.

Beyond accuracy, DAIST's compositional structure also reduces training overhead. When a new topology is introduced, only novel block types need to be trained, while existing block models are frozen and reused at zero additional cost.

\subsection{Cross-Topology Validation: Interposer--MC--TIM}
\label{subsec:cross_topology}

As established in Section~\ref{subsec:experiment_setup}, the interposer block carries no
internal heat sources in either system configuration, meaning it learns only
conduction-driven thermal transport rather than power-driven transport.
This property motivates a direct test of compositional reusability. A DAIST
instance for the advanced packaging system (Interposer--MC--TIM) is assembled using the interposer
block model trained exclusively on multi-chiplet (Interposer--SoC--HBM) data. The interposer block model stays frozen
without any fine-tuning.
The MC and TIM block models are each trained independently on their own separately
generated 5,000-case datasets with an 8-1-1 split.
Table~\ref{tab:cross_topology_reuse} summarizes the training and testing configuration
for each block, confirming that the interposer model is evaluated on a stack configuration unseen during training.

\begin{table}[t]
  \centering
  \caption{Cross-topology block reuse: training and testing configurations.
    The interposer model is frozen from the multi-chiplet system and receives
    no MC--TIM training data.}
  \label{tab:cross_topology_reuse}
  \small
  \begin{tabular}{@{}lll@{}}
    \toprule
    \textbf{Block} & \textbf{Trained on} & \textbf{Tested in} \\
    \midrule
    Interposer & SoC--HBM (frozen) & MC--TIM \\
    MC         & MC--TIM            & MC--TIM \\
    TIM        & MC--TIM            & MC--TIM \\
    \bottomrule
  \end{tabular}
\end{table}

\begin{figure*}[t]
  \centering
  \includegraphics[width=\textwidth]{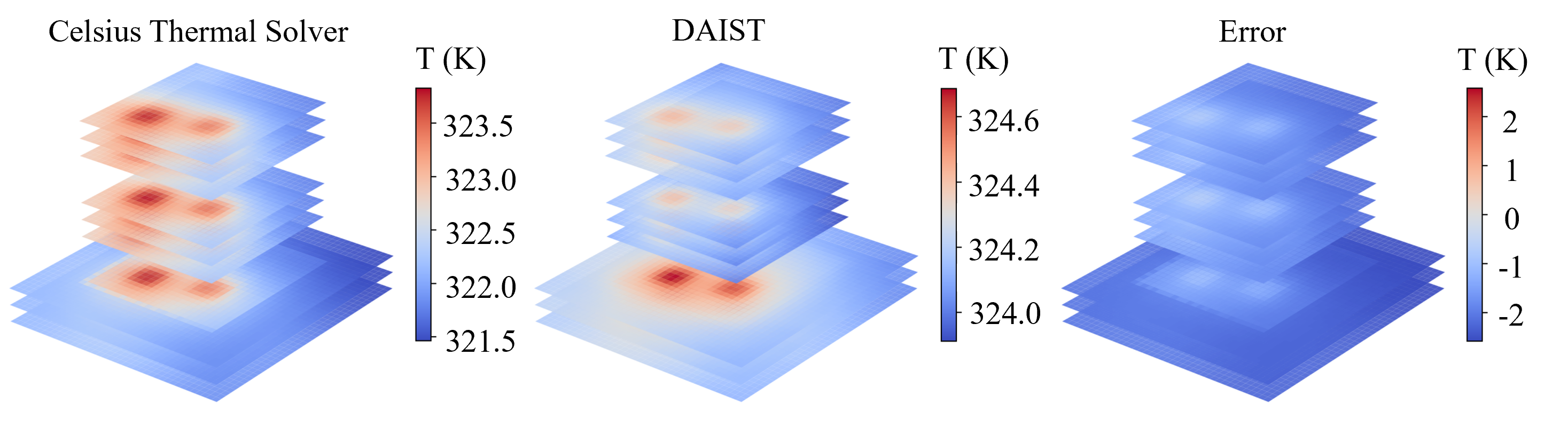}
  \caption{Three-dimensional thermal visualization of the advanced packaging assembly
    (Interposer--MC--TIM) with a frozen interposer model transferred from the
    multi-chiplet system.
    Left: Celsius Thermal Solver reference. Center: DAIST prediction. Right: pointwise error.}
  \label{fig:advanced_package_3d}
\end{figure*}

Figure~\ref{fig:advanced_package_3d} shows the three-dimensional temperature field for a
representative test case of the advanced packaging system.
The Celsius reference exhibits two hotspots near the left corner,
with the interposer layer alone accounting for a temperature range of approximately
2\,K.
DAIST captures the same spatial hotspot pattern but with two notable discrepancies.
First, DAIST shows a compressed thermal gradient (under 1\,K vs. 2\,K).
Second, the ground truth exhibits a sharp temperature discontinuity at the
interposer--MC interface between the MC and exposed regions, which DAIST
fails to reproduce.
Both discrepancies are concentrated in the interposer layer, with errors reaching up to
$\pm 2$\,K.

These discrepancies are expected.
In the multi-chiplet system, the interposer receives heat from multiple discrete die
footprints with independently sampled power maps, producing a distributed thermal load
across its surface. By contrast, the advanced packaging system concentrates heat into 1–4 small regions within a single MC block, creating a much sharper, more distinct flux pattern.
The frozen interposer block has never encountered this flux distribution, which accounts
for both the compressed thermal gradient and the missing temperature discontinuity at
the MC boundary. Closing this domain gap through lightweight fine-tuning on a small number of advanced
packaging samples is a natural direction for future work.

Validated on 100 held-out MC--TIM test cases, DAIST achieves an MAE of $1.056$\,K, RMSE of $1.135$\,K,
MAPE of $3.23\times10^{-3}$, and a mean per-case
maximum absolute error of $2.396$\,K. These cases were entirely unseen by the interposer model during training.
While accuracy decreases relative to the in-distribution multi-chiplet result ($0.253$\,K MAE), 
this reflects the cost of zero-retraining cross-topology reuse. Specifically, the interposer model
generalizes to a fundamentally different stack (from multiple chiplets to a single MC)
without any exposure to MC--TIM training data.

\smallskip\noindent\textbf{Convergence and Speedup.}
With a predefined convergence criterion of 0.1\%, 
DAIST requires on average $15.3$ interface iterations to
converge for the three-block sequential chain (interposer $\to$ MC $\to$ TIM).
This higher iteration count relative to the multi-chiplet case ($5.2$ iterations)
is due to the sequential topology of the assembly. The interposer has no direct
interface with the TIM block, so thermal information must propagate through an
intermediate MC block before the full chain reaches consistency.
In a parallel coupling graph (as in the multi-chiplet system, where all dies interface
directly with the interposer), interface inconsistencies are corrected in fewer global
passes. However, in a linear chain, each pass can only propagate information one hop further,
requiring more iterations.
Table~\ref{tab:mctim_iter_mae} shows the full iteration-wise 
breakdown.
The reference solve for the full advanced packaging system takes $197$\,s.
DAIST converges in approximately $2$\,s, yielding a speedup of $\mathbf{99\times}$.

\begin{table}[t]
  \centering
  \caption{Mean MAE and wall time at fixed iteration counts
for advanced packaging test set.}
  \label{tab:mctim_iter_mae}
  \small
  \begin{tabular}{@{}ccc@{}}
    \toprule
    \textbf{Iterations} & \textbf{Mean MAE (K)} & \textbf{Wall time (s)} \\
    \midrule
    $1$  & $33.282$  & $0.328$ \\
    $5$  & $5.657$  & $0.796$ \\
    $10$  & $1.914$  & $1.381$ \\
    $15$ & $1.239$ & $1.966$ \\
    $20$ & $0.997$ & $2.551$ \\
    \midrule
    Celsius reference & --- & $197$ \\
    \bottomrule
  \end{tabular}
\end{table}

\section{Conclusion}
\label{sec:conclusion}

We presented DAIST, a composable thermal solver that decomposes 
heterogeneous 2.5D/3D-IC packages into modular blocks, replaces 
subdomain solvers with neural operators, and couples them through 
iterative interface exchange of temperature and heat flux. 
Evaluated on a multi-chiplet system and an advanced packaging system, 
DAIST achieves up to 178× speedup over Celsius with mean temperature 
errors of 0.068\% and 0.323\%, respectively. 
Notably, the advanced packaging result is achieved with a frozen 
interposer model transferred directly from the multi-chiplet system, 
demonstrating cross-topology reuse of block-level models without retraining. 
Future work will explore transfer learning to narrow the cross-topology 
accuracy gap, extend the framework to more complex 3D-IC packages, 
and apply DAIST to thermal-aware design optimization.

\bibliographystyle{IEEEtran}
\bibliography{refs}

\end{document}